\documentclass{article}

\PassOptionsToPackage{numbers, compress}{natbib}
 \usepackage[preprint]{neurips_2026}

\usepackage[utf8]{inputenc} 
\usepackage[T1]{fontenc}    
\usepackage{hyperref}       
\usepackage{url}            
\usepackage{booktabs}       
\usepackage{amsfonts}       
\usepackage{nicefrac}       
\usepackage{microtype}      
\usepackage{xcolor}         

\usepackage{multirow}
\usepackage{amsmath}
\usepackage{graphicx} 
\usepackage{xspace}

\usepackage{array}
\usepackage{siunitx}
\usepackage{adjustbox}
\usepackage{subcaption}
\usepackage{wrapfig}
\usepackage{comment}
\usepackage{amssymb}
\usepackage{enumitem}
\usepackage{stackrel}
\usepackage{wrapfig}
\usepackage{float}

\newcommand{\Method}{CrystalBoltz\xspace}
\newcommand{\ci}[1]{\,{\scriptstyle\pm\,#1}}

\title{\Method: End-to-End Protein Structure Determination via Experiment-Guided Diffusion \\ for X-Ray Crystallography}

\author{%
  Minseo Kim \\
  Stanford University
  \And
  Huanghao Mai\thanks{Equal contribution} \\
  SLAC National Accelerator Laboratory
  \And
  Jay Shenoy\footnotemark[1] \\
  Stanford University
  \AND
  Alec Follmer \\
  UC Davis
  \And
  Gordon Wetzstein \\
  Stanford University
  \And
  Frédéric Poitevin \\
  SLAC National Accelerator Laboratory
}

\begin{document}

\maketitle

\begin{abstract}
  Generative models trained on public databases of protein structures, most of which have been determined by X-ray crystallography, now provide powerful priors for structure prediction. However, they are not readily conditioned on the measurements from a new crystallographic experiment, limiting their use for X-ray structure determination. In crystallography, the measured structure-factor amplitudes do not by themselves determine an electron density map or atomic structure because the associated phases are unobserved and must be inferred. Structure determination therefore remains an inverse problem in which candidate models must be both structurally plausible and consistent with measured diffraction data, often requiring substantial manual refinement by human experts. Emerging methods aim to incorporate experimental information more directly into predictive and refinement workflows. We present CrystalBoltz, a generative framework that casts crystallographic refinement as Bayesian inference over atomic structures and operates directly on structure-factor amplitudes. CrystalBoltz moves from unguided generation with a pre-trained prior over protein structures to experiment-guided posterior sampling, followed by atomic coordinate and B-factor refinement. Across multiple protein crystallography datasets, CrystalBoltz attains lower coordinate RMSD and lower R-factors than the strongest baselines considered, while reducing runtime by a factor of 33 relative to existing experimentally guided refinement.
\end{abstract}

\section{Introduction}
Protein structure predictors such as AlphaFold provide strong sequence-conditioned priors~\citep{jumper2021,alphafold3}, but experimental structure determination is ultimately constrained by agreement with measured data.
This tension is especially important in X-ray crystallography, the predominant technique for solving protein structures, where measured structure-factor amplitudes do not include phases and therefore do not directly determine an atomic structure~\citep{taylor2003phase,hendrickson2023phaseproblem}. Aligning structure predictors with crystallographic observables is therefore an active area of research, particularly as higher experimental throughput increases the need for scalable computational approaches. Generative structure predictors create new opportunities to bridge learned structural priors and experimental measurements at lower computational cost.
Here, we present CrystalBoltz, which efficiently and consistently improves the agreement of generative predictions with crystallographic measurements, namely structure-factor amplitudes, and enables end-to-end model building and refinement using differentiable likelihoods.


\Method formulates structure determination as diffusion posterior sampling~\citep{dps}, a general strategy for combining generative priors with likelihood gradients for noisy inverse problems, directly linking experimental structure-factor data with the atomic structure of proteins in an end-to-end fashion. \Method couples (i) experiment-guided diffusion updates using crystallographic likelihood terms with (ii) a post-guidance refinement stage that optimizes coordinates and B-factors against experimental targets. This design integrates learned structural priors with crystallography-native objectives in one inference pipeline.

Empirically, on experimental structure-factor datasets, \Method improves both structural and crystallographic agreement relative to strong baselines while substantially reducing runtime relative to state-of-the-art experimentally guided refinement~\citep{fadini2025rocket} in our setup. These results suggest that moving experimental conditioning into generative sampling is an effective path toward faster, data-consistent structure determination.

\paragraph{Contributions.} \begin{itemize} \item We cast crystallographic structure determination as experiment-guided posterior sampling with a diffusion prior. \item We derive crystallography-native likelihood guidance from differentiable structure-factor objectives, enabling the model to resolve \textbf{large conformational changes}. \item We combine guided sampling with post-guidance crystallographic refinement of coordinates and B-factors. \item We achieve better agreement metrics at \textbf{$33.3\times$ lower runtime} than the state of the art on six experimental targets. \end{itemize}
\section{Background}

\subsection{X-Ray Crystallography Structure Determination and Forward Model}

Given a protein structure defined by its set of constitutive atoms and their coordinates $\vec{x}_j$, an X-ray crystallography experiment probes the structure by measuring the structure factors, which refer to the scattering at different spatial frequency $\vec{h}$, also known as the Miller index~\citep{rhodes2006crystallography,rupp2009biomolecular}. Upon measurement, the phases of structure factors are lost, and the observed structure factor amplitudes $|\mathbf{F}_o|$ are obtained through a data-reduction pipeline that includes diffraction-spot finding, integration, scaling and merging of symmetry-equivalent measurements, as well as uncertainty estimation~\citep{kabsch2010xds,winter2010xia2,winter2018dials}. While there are experimental techniques to provide an initial estimation of phases~\citep{hendrickson2023phaseproblem,rupp2009biomolecular}, molecular replacement using similar structures has been the predominant method due to its efficiency and capability to leverage AlphaFold predictions ~\citep{keegan2024success,oeffner2022putting}. Finally, phases are updated through refining the atomic structure, yielding an electron density map when combined with the observed amplitudes.

Crystallographic structural determination and refinement seek to achieve satisfying agreement between the calculated structure factor amplitudes $|\mathbf{F}_c|$ and the observed ones $|\mathbf{F}_o|$. 


A forward model for the calculated structures in macromolecular crystallography typically considers two scattering sources -- the protein of interest and the bulk solvent in the protein crystal~\citep{afonine2013bulk}:
\begin{equation}
    \mathbf{F}_c(\vec{h}) = k_\text{total} (\mathbf{F}_{\text{protein}}+k_{\text{mask}}\cdot \mathbf{F}_{\text{solvent}})
\end{equation}



The protein contribution, following the notation in \citep{sfcalculator}, is computed as 
\begin{equation}
\mathbf{F}_\text{protein}(\vec{h}) = \sum_{G} \sum_{j} f_j(\vec{h}) \exp[-2\pi \mathbf{i}(\vec{h}\cdot G(\vec{x}_j))] \cdot \text{DWF}_j(\vec{h}) \cdot O_j
\end{equation}
That sums the contribution of each atom $j$ -- approximated by the atomic scattering factors $f_j$ -- with occupancy $O_j$ at the real-space position $\vec{x}_j$ as well as the equivalent positions related by the symmetry operation $G$. The resulting structure factor is then attenuated by 
the Debye--Waller factor $\text{DWF}_j(\vec{h})$ which is commonly modelled as a standard Gaussian, \emph{e.g.} $\exp\left(-B_j h^{2}/4\right)$ in the isotropic case, parameterized by the atomic B-factor $B_j$. The contribution of bulk solvent scattering $\mathbf{F}_{\text{solvent}}$ 
is conventionally calculated from a space-filling mask that occupies the remaining volume of the crystal's unit cell not occupied by protein atoms~\citep{afonine2013bulk}. Because this masking procedure prevents differentiable calculation of $\mathbf{F}_{\text{solvent}}$, we leverage a new tool named \texttt{SFCalculator} that proposes a differentiable method for bulk solvent ~\citep{sfcalculator}. 
To put the calculated structure factors on a comparable scale with the observed values, the protein and the solvent scattering effects are combined using a weighted sum governed by scaling factors 
$k_{\text{total}}$ and $k_{\text{mask}}$ (see Appendix~\ref{hyperparam} for more details).
As the phase information of the Fourier signals is lost upon experimental measurement, the final step of the forward model takes only the amplitude of the structure factors $|\mathbf{F}_c|$. 


\subsection{Related work}
CrystalBoltz builds on three lines of work. The first is conventional macromolecular crystallographic refinement and model building. Software such as REFMAC5 and Phenix optimizes atomic coordinates, B-factors, scale terms, solvent parameters, and stereochemical restraints against diffraction-derived targets~\citep{murshudov1997,refmac5,phenix2019,afonine2012towards,afonine2013bulk}. These workflows remain central to crystallographic structure determination, but they generally require an initial model close enough for molecular replacement, model building, and local refinement to succeed.

The second line of work uses protein structure predictors to address this starting-model requirement and to explore alternatives to a single default prediction. AlphaFold2 showed that sequence-conditioned models can produce highly accurate structures from sequence information alone~\citep{jumper2021}, and AlphaFold predictions have expanded the set of usable starting models for phasing and refinement~\citep{terwilliger2023,oeffner2022putting,mccoy2022alphafold_phasing,keegan2024success,wang2025alphafold_guided_mr}. Subsequent work has altered multiple-sequence alignments or sequence profiles to sample alternative conformations from the AlphaFold2 pipeline~\citep{del2022sampling,stein2022speach_af,wayment2024predicting,kalakoti2025afsample2}. ROCKET extends this representation-level strategy by optimizing latent multiple-sequence alignment representations in AlphaFold2/OpenFold against experimental signals, including X-ray diffraction data~\citep{fadini2025rocket,ahdritz2024openfold}. These methods demonstrate that learned predictors can be steered beyond a single default prediction, but they do so through sequence input or latent-representation optimization rather than direct generative conditioning on observed structure-factor amplitudes.

The third line of work concerns diffusion-based structure predictors and their use as priors for inverse problems. Recent predictors such as AlphaFold3, Boltz-2, RF3, and Protenix use diffusion-based generative modeling, in which sampling proceeds through intermediate denoising states~\citep{alphafold3,boltz2,rf3,protenix}. This sampling trajectory provides a natural point of intervention for experimental conditioning. Diffusion Posterior Sampling (DPS) formalizes this idea by augmenting a pretrained diffusion prior with likelihood gradients from a forward model~\citep{dps}. Related diffusion-prior methods have been applied to broader inverse problems~\citep{survey_posterior_sampling}, including protein-space inverse problems~\citep{levy2024protein_space_inverse}. In structural biology, experiment-guided AlphaFold formulations infer conformational ensembles from measurement constraints~\citep{maddipatla2025inverse}, CryoBoltz guides Boltz-1 sampling with cryo-EM density maps~\citep{raghu2025cryoboltz,boltz1}, and EmbedOpt guides Protenix through sequence-embedding optimization against cryo-EM density maps~\citep{li2026robust,protenix}. CrystalBoltz follows this inference-time conditioning principle, but uses Boltz-2 as the sequence-conditioned prior~\citep{boltz2}, guides sampling with crystallographic structure-factor amplitudes, and then refines coordinates and B-factors against crystallographic targets.

\section{Methods}

\Method has two stages: experiment guidance during diffusion sampling and post-guidance refinement. First, we sample structures from a sequence-conditioned diffusion prior while applying gradients from a differentiable crystallographic likelihood on observed structure-factor amplitudes. Second, we refine the denoised coordinates and isotropic B-factors against crystallographic objectives. This separates global, prior-guided conformational search from local crystallographic fitting.

\begin{figure}[h]
    \centering
    \includegraphics[width=\linewidth]{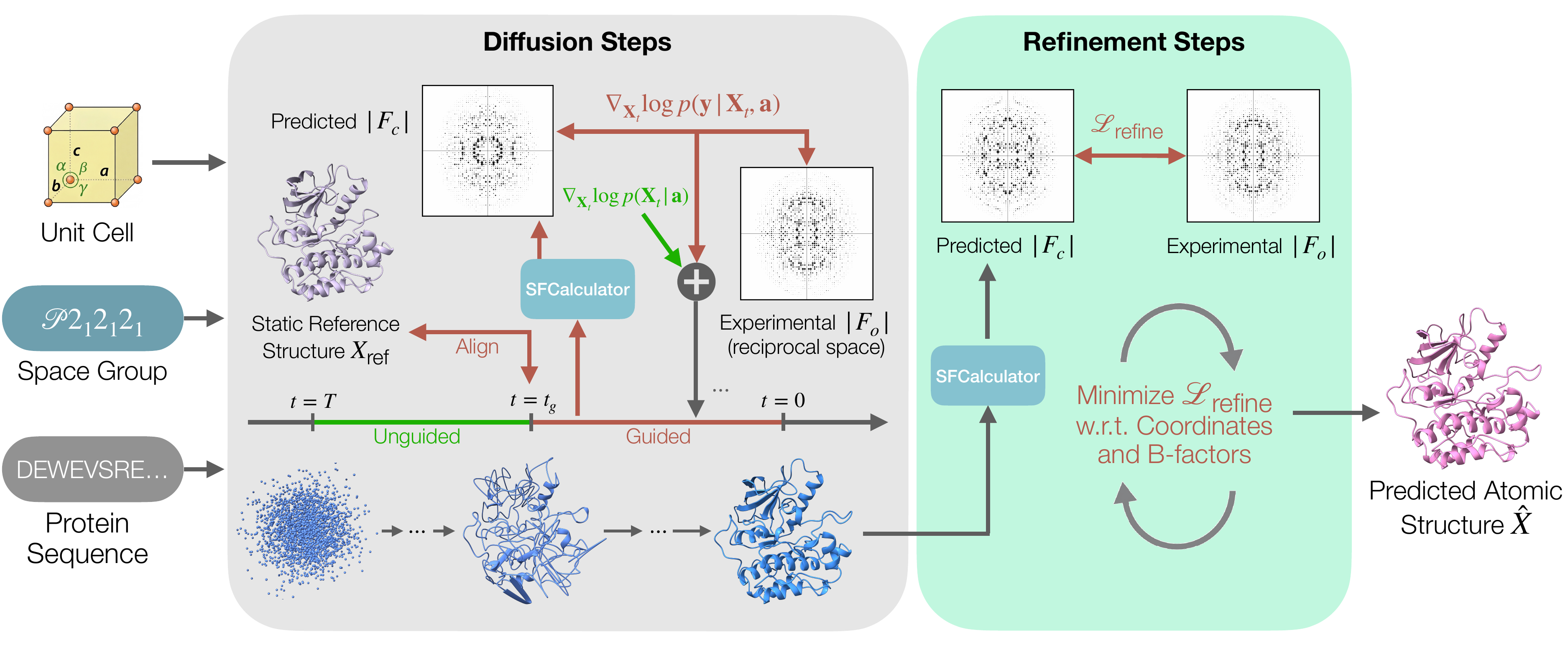}
    \caption{\textbf{Algorithm Overview.} \Method has two phases. Phase 1 runs Boltz diffusion conditioned on the protein sequence and known crystallographic parameters (unit cell and space group). Sampling begins unguided; once a coarse structure has formed, experimental guidance is switched on at step $t_g$ and continues to $t=0$. Phase 2 takes the resulting structure and refines atomic coordinates and B-factors against the experimental data.}
    \label{fig:placeholder}
\end{figure}

\subsection{Diffusion Posterior Sampling of Protein Structures}
\label{dps_section}
Given a protein sequence $\mathbf{a}$, crystal parameters $\mathbf{c} := (\mathbf{u}, \mathcal{G})$ comprising the unit cell $\mathbf{u}$ and space group $\mathcal{G}$, and experimental observables $\mathbf{y}:=|\mathbf{F}_o|$, we sample a structure consisting of $m$ atoms as $\mathbf{X}=(\vec{x}_0, \ldots, \vec{x}_m)$ from the posterior distribution:
\begin{equation}
    p(\mathbf{X}|\mathbf{a},\mathbf{c},\mathbf{y}) \propto p(\mathbf{y}|\mathbf{X},\mathbf{a},\mathbf{c}) \cdot p(\mathbf{X}|\mathbf{a})
    \label{bayes}
\end{equation}

where $p(\mathbf{X}|\mathbf{a}) = p(\mathbf{X}|\mathbf{a}, \mathbf{c})$ represents the structural prior from Boltz-2 and $p(\mathbf{y}|\mathbf{X},\mathbf{a},\mathbf{c})$ encodes the experimental likelihood through the nonlinear forward operator.

Boltz-2 uses a variance-preserving diffusion model for structure generation, where the reverse stochastic differential equation (SDE) is defined as:
\begin{equation}
    d\mathbf{X}_t = -\left(\frac{1}{2}\mathbf{X}_t \, dt + \nabla_{\mathbf{X}_t} \log p_t(\mathbf{X}_t|\mathbf{a})\right)\beta_t \, dt + \sqrt{\beta_t} \, d\mathbf{W}
    \label{reverse_sde}
\end{equation}

where $\beta_t$ defines the noise schedule, $d\mathbf{W}$ represents Brownian motion, and $\nabla_{\mathbf{X}_t} \log p_t(\mathbf{X}_t|\mathbf{a})$ is the score function modeled by AlphaFold3's denoising network at time $t$. We can convert the unconditional sampling in Eq.~\ref{reverse_sde} into conditional sampling by injecting Eq.~\ref{bayes} into Eq.~\ref{reverse_sde}:
\begin{equation}
    d\mathbf{X}_t = -\left[\frac{1}{2}\mathbf{X}_t \, dt + \left(\nabla_{\mathbf{X}_t} \log p_t(\mathbf{X}_t|\mathbf{a}) + \nabla_{\mathbf{X}_t} \log p_t(\mathbf{y}|\mathbf{X}_t,\mathbf{a},\mathbf{c})\right)\right]\beta_t \, dt + \sqrt{\beta_t} \, d\mathbf{W}
    \label{reverse_sde_cond}
\end{equation}


This allows us to effectively sample from the posterior by incorporating the experimental guidance. 

The conditional density $p(\mathbf{y}|\mathbf{X}_t,\mathbf{a},\mathbf{c})$ in Eq.~\ref{reverse_sde_cond} is intractable, since it requires marginalizing the crystallographic forward model over the posterior $p(\mathbf{X}_0|\mathbf{X}_t, \mathbf{a})$ of clean structures. Following  DPS~\cite{dps}, we apply Tweedie's formula and approximate the marginal by a point estimate at the posterior mean,
\begin{equation}
    p(\mathbf{y}|\mathbf{X}_t, \mathbf{a},\mathbf{c}) \;\approx\; p\!\left(\mathbf{y}\mid \hat{\mathbf{X}}_0(\mathbf{X}_t, \mathbf{a}),\mathbf{c}\right), \qquad \hat{\mathbf{X}}_0(\mathbf{X}_t, \mathbf{a}) \,:=\, \mathbb{E}[\mathbf{X}_0|\mathbf{X}_t, \mathbf{a}],
\end{equation}
which the Boltz-2 denoiser supplies directly as its one-step prediction. 

\paragraph{Likelihood function.}
The differentiability of the calculated structure factors $\mathbf{F}_c(\vec{h})$ via \texttt{SFCalculator}~\cite{sfcalculator} enables physics-informed, gradient-based likelihood guidance during sampling. We compare the sampled structure $\mathbf{X}_t$ against experimental amplitudes at each Miller index $\vec{h}$, working with normalized amplitudes $|\mathbf{E}_c(\vec{h})|$ and $|\mathbf{E}_o(\vec{h})|$ such that $\mathbb{E}[|\mathbf{E}_c(\vec{h})|^2] = \mathbb{E}[|\mathbf{E}_o(\vec{h})|^2] = 1$, and consider two likelihood targets.

The first is a Gaussian error model, which yields a heteroscedastic least-squares loss weighted by the per-reflection uncertainty $\tilde{\sigma}_{\vec{h}}$:
\begin{equation}
    L_{\text{gauss}}(\mathbf{X}_t) = \sum_{\vec{h}} \frac{\big(|\mathbf{E}_o(\vec{h})| - |\mathbf{E}_c(\vec{h})|\big)^2}{2\tilde{\sigma}_{\vec{h}}^2}.
\end{equation}

The second is the Rice distribution, a more sophisticated target in macromolecular crystallography. By the central limit theorem, structure factors are complex normal for acentric reflections and real normal for centric reflections~\citep{srinivasan1976some}; marginalizing the unmeasured phase yields a Rice density, with the parameter $\sigma_A$ additionally absorbing atomic coordinate error~\citep{read1990rice,read2016log}. Splitting reflections into acentric ($A$) and centric ($C$) sets (see Appendix~\ref{rice_dist} for details on how its defined):
\begin{equation}
    L_{\text{rice}}(\mathbf{X}_t) = - \left[ \sum_{\vec{h} \in A} \log p_a \big(|\mathbf{E}_o(\vec{h})|;\, |\mathbf{E}_c(\vec{h})|\big) + \sum_{\vec{h} \in C} \log p_c \big(|\mathbf{E}_o(\vec{h})|;\, |\mathbf{E}_c(\vec{h})|\big) \right].
    \label{eq:rice_loss}
\end{equation}

The combined guidance signal is
\begin{equation}
    \nabla_{\mathbf{X}_t} \log p(\mathbf{y}|\mathbf{X}_t, \mathbf{a}, \mathbf{c}) = - \rho \nabla_{\mathbf{X}_t} \big( \lambda_{\text{gauss}} L_{\text{gauss}}(\mathbf{X}_t) + \lambda_{\text{rice}} L_{\text{rice}}(\mathbf{X}_t) \big),
    \label{log_likeli}
\end{equation}
with step size $\rho$ and empirically chosen weights $\lambda_{\text{gauss}}, \lambda_{\text{rice}}$.

In practice, $L_{\text{gauss}}$ and $L_{\text{rice}}$ are evaluated on $\hat{\mathbf{X}}_0$, and the gradient is backpropagated through the denoiser to $\mathbf{X}_t$. We align the denoiser prediction $\hat{\mathbf{X}}_0$ to the crystal frame using a static reference structure $\mathbf{X}_{\text{ref}}$ before it is passed to \texttt{SFCalculator} (see Appendix~\ref{rigid_body_supp} for more details). This step is essential as the generative structure prior outputs prediction in an arbitrary frame, but the protein structure can only be meaningfully compared with the experimental observable when the \textit{crystal} structure in the forward model is consistent -- that is the protein configuration in the unit cell frame.

\subsection{Post-Guidance Refinement}
\label{post-hoc_refine}
DPS guidance improves the global structure and lowers RMSD, but operates at low resolution and does not resolve local details such as side-chain rotamers or B-factors, following a similar practice in ROCKET\citep{fadini2025rocket}. A post-guidance refinement step against the experimental data is therefore required to correct these local errors and obtain accurate R-factors. The diffusion sampling stage produces a predicted structure $\hat{\mathbf{X}}_0$, which we align to crystal-frame coordinates before refinement. Each DPS step requires backpropagation through the full denoiser Jacobian $\partial\hat{\mathbf{X}}_0/\partial\mathbf{X}_t$; refinement instead acts on the final denoised structure and requires only the crystallographic forward model.

We jointly optimize atomic coordinates and B-factors $\mathbf{B} = (b_1, \ldots, b_m)$:
\begin{equation}
    (\mathbf{X}^*, \mathbf{B}^*) = \underset{\mathbf{X},\; b_j \in [1,\,80]\,\text{\AA}^2}{\arg\min}\;\; L_{\mathrm{refine}}(\mathbf{X}, \mathbf{B}),
    \label{eq:refine}
\end{equation}
initialized at $(\hat{\mathbf{X}}_0,\, \mathbf{B}_0)$ and run for $T_{\mathrm{refine}}$ steps of Adam. Whereas the DPS stage requires a likelihood-based objective to guide a stochastic process, the refinement objective is unconstrained in this respect. We exploit this flexibility by optimizing the crystallographic evaluation metric directly, either the correlation coefficient $\mathrm{CC}$ or the R-factor, so that the objective minimized during refinement coincides with the conventionally reported metrics in crystallography. The R-factor is reported as $R_\mathrm{work}$ on the working reflections and $R_\mathrm{free}$ on a held-out reflection set, following the cross-validation principle introduced by Br{\"u}nger to reduce overfitting during macromolecular refinement~\citep{brunger1992free}. B-factors enter the forward model through the Debye--Waller envelope $\exp(-b_j|\vec{h}|^2/4)$. They are initialized from Boltz-2 pLDDT scores using the conversion of~\cite{baek2021} and clamped to $[1, 80]$~\AA$^2$ after each gradient step to enforce physical validity.

\section{Experiments}

\begin{figure}[h]
\centering
\includegraphics[width=\linewidth]{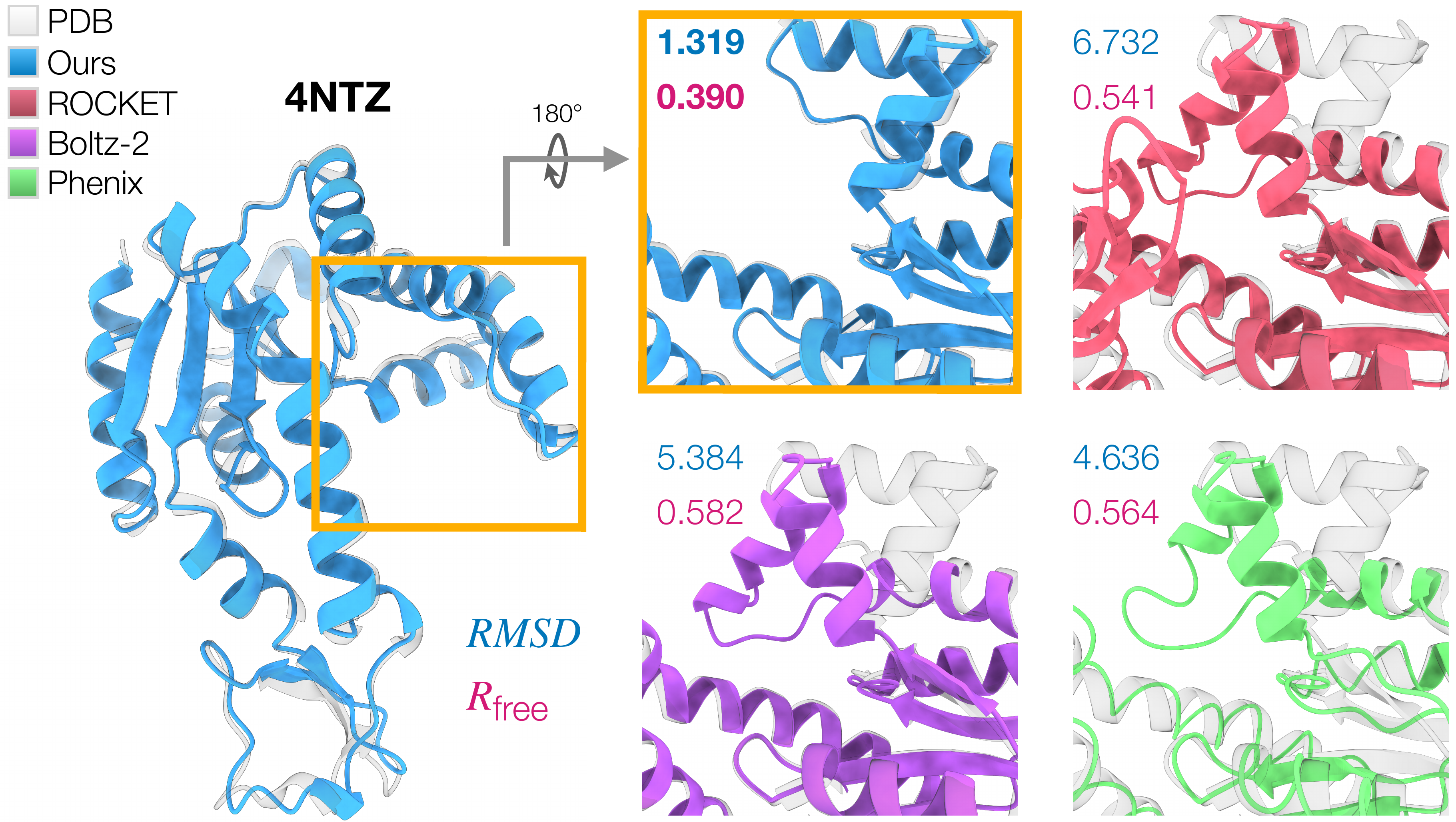}
\caption{\textbf{Qualitative results on PDB 4NTZ.} \Method can correct large conformation change while other baselines cannot. Not only is there a significant improvement on the global RMSD, the R-factor is also reduced.}
\label{fig:4ntz_qual_fig}
\end{figure}

\paragraph{Datasets and metrics.} We evaluate our method on six proteins from the PDB~\cite{pdb}:
8DWN, 4NTZ, 7O51, 7SEZ, 7VNX, and 1L63. The set spans diffraction
resolutions of 1.69--2.20\AA\, and protein sizes of 164--306 residues, covering a range of folds and difficulty
levels. Currently, we limit our choice of proteins to single chain in the asymmetric unit for the convenience of alignment-based evaluation. As we establish proof-of-concept on crystallographic guidance of generative prior in this work,  we expect multi-chain extension a natural follow-up on the engineering side, especially since  Boltz-2 natively predicts multi-chain complexes and the forward model is agnostic to chain counts. For each
PDB entry, we use the deposited experimental structure factor amplitudes as the supervision target, and we inherit
the deposited free-flag set to define the held-out
$R_\text{free}$ measurements, ensuring that the performance metrics are comparable
across all evaluated methods. We report four standard metrics.
\textbf{Global RMSD} and \textbf{C$\alpha$ RMSD} (\AA, lower is better)
measure all-atom and backbone agreement against the deposited PDB
coordinates. The
crystallographic \textbf{$\textbf{R}$-factors} $R_\text{work}$ and
$R_\text{free}$ (lower is better)~\cite{brunger1992free}
measure normalized L1 agreement between calculated and experimental
amplitudes on the working and held-out ($\sim$5\%) reflection sets,
respectively, with $R_\text{free}$ serving as the standard
cross-validated metric of model fit. Together, RMSD captures geometric
fidelity to the deposited structure, while $R_\text{work}/R_\text{free}$
capture fidelity to the experimental observations.

\paragraph{Baselines.} We compare \Method against four baselines spanning purely
physics-based, purely data-driven, and data-guided methods.
\textbf{PDB} reports $R_\text{work}$ and $R_\text{free}$ of the
deposited structure as released by the original depositors and serves
as a reference for crystallographic quality rather than a competing
prediction; RMSD is omitted as the deposited structure is the reference
against which all RMSDs are computed.
\textbf{Phenix}~\cite{phenix2019} is the standard refinement
package in macromolecular crystallography. We initialize
\texttt{phenix.refine} from the Boltz-2 prediction and run its default
protocol against the experimental amplitudes, giving a strong classical
baseline against which the contribution of a generative prior can be
measured. \textbf{Boltz-2}~\cite{boltz2} is the unguided structure prediction
model that supplies the prior in our method; it receives no
experimental input and serves as the unconditioned baseline.
\textbf{ROCKET}~\cite{fadini2025rocket} is a recent data-guided method that
conditions AlphaFold2/OpenFold~\cite{jumper2021,ahdritz2024openfold} predictions on diffraction data by
optimizing the MSA embeddings using similar likelihood targets. It is the most directly comparable
baseline to our approach, as both methods steer a pretrained protein structure predictor with experimentally measured structure factor amplitudes.

\begin{table}[h!]
\centering
\caption{\textbf{Experimental $|\mathbf{F}_o|$ evaluation results.} CrystalBoltz achieves the best RMSD and R-factors on the majority of targets and improves unguided Boltz-2 predictions, particularly on R-factors. Each value reports the mean across 3 best samples out of 20 samples per protein with the corresponding 95\% confidence interval. \textbf{Bold} indicates the statistically significant best value(s); ties with overlapping 95\% confidence intervals are bolded together.}
\vspace{0.5em}
\resizebox{\textwidth}{!}{%
\begin{tabular}{l|l|c|cccc}
\toprule
\textbf{Structure} & \textbf{Metrics} & \textbf{PDB} & \textbf{Phenix} & \textbf{Boltz-2} & \textbf{ROCKET} & \textbf{CrystalBoltz} \\
\midrule
\multirow{4}{*}{8DWN} & RMSD (\AA, $\downarrow$) & $-$ & $2.766\ci{0.328}$ & $2.653\ci{0.281}$ & $2.202\ci{0.216}$ & $\mathbf{1.319}\ci{\mathbf{0.143}}$ \\
 & C$\alpha$ RMSD (\AA, $\downarrow$) & $-$ & $2.499\ci{0.319}$ & $2.457\ci{0.265}$ & $2.031\ci{0.196}$ & $\mathbf{0.828}\ci{\mathbf{0.119}}$ \\
 & $R_\mathrm{work}$ ($\downarrow$) & 0.190 & $0.519\ci{0.016}$ & $0.472\ci{0.012}$ & $0.376\ci{0.007}$ & $\mathbf{0.338}\ci{\mathbf{0.029}}$ \\
 & $R_\mathrm{free}$ ($\downarrow$) & 0.250 & $0.557\ci{0.010}$ & $0.474\ci{0.008}$ & $0.382\ci{0.009}$ & $\mathbf{0.337}\ci{\mathbf{0.031}}$ \\
\midrule
\multirow{4}{*}{4NTZ} & RMSD (\AA, $\downarrow$) & $-$ & $4.636\ci{1.787}$ & $4.538\ci{1.800}$ & $8.772\ci{2.905}$ & $\mathbf{1.299}\ci{\mathbf{0.364}}$ \\
 & C$\alpha$ RMSD (\AA, $\downarrow$) & $-$ & $4.324\ci{1.777}$ & $4.276\ci{1.798}$ & $8.352\ci{2.778}$ & $\mathbf{0.769}\ci{\mathbf{0.332}}$ \\
 & $R_\mathrm{work}$ ($\downarrow$) & 0.184 & $0.530\ci{0.027}$ & $0.519\ci{0.017}$ & $\mathbf{0.491}\ci{\mathbf{0.036}}$ & $\mathbf{0.447}\ci{\mathbf{0.038}}$ \\
 & $R_\mathrm{free}$ ($\downarrow$) & 0.216 & $0.545\ci{0.012}$ & $0.570\ci{0.011}$ & $0.554\ci{0.029}$ & $\mathbf{0.483}\ci{\mathbf{0.036}}$ \\
\midrule
\multirow{4}{*}{7O51} & RMSD (\AA, $\downarrow$) & $-$ & $1.556\ci{0.112}$ & $0.732\ci{0.018}$ & $1.125\ci{0.043}$ & $\mathbf{0.651}\ci{\mathbf{0.031}}$ \\
 & C$\alpha$ RMSD (\AA, $\downarrow$) & $-$ & $1.202\ci{0.071}$ & $0.282\ci{0.025}$ & $0.610\ci{0.038}$ & $\mathbf{0.224}\ci{\mathbf{0.029}}$ \\
 & $R_\mathrm{work}$ ($\downarrow$) & 0.155 & $0.577\ci{0.002}$ & $0.582\ci{0.006}$ & $0.341\ci{0.009}$ & $\mathbf{0.258}\ci{\mathbf{0.019}}$ \\
 & $R_\mathrm{free}$ ($\downarrow$) & 0.200 & $0.651\ci{0.014}$ & $0.626\ci{0.012}$ & $0.381\ci{0.007}$ & $\mathbf{0.278}\ci{\mathbf{0.025}}$ \\
\midrule
\multirow{4}{*}{7SEZ} & RMSD (\AA, $\downarrow$) & $-$ & $1.964\ci{0.452}$ & $1.108\ci{0.039}$ & $2.127\ci{0.184}$ & $\mathbf{1.014}\ci{\mathbf{0.038}}$ \\
 & C$\alpha$ RMSD (\AA, $\downarrow$) & $-$ & $1.369\ci{0.522}$ & $0.691\ci{0.083}$ & $1.509\ci{0.216}$ & $\mathbf{0.506}\ci{\mathbf{0.089}}$ \\
 & $R_\mathrm{work}$ ($\downarrow$) & 0.197 & $0.557\ci{0.005}$ & $0.571\ci{0.002}$ & $0.412\ci{0.007}$ & $\mathbf{0.348}\ci{\mathbf{0.022}}$ \\
 & $R_\mathrm{free}$ ($\downarrow$) & 0.228 & $0.567\ci{0.008}$ & $0.591\ci{0.004}$ & $0.451\ci{0.009}$ & $\mathbf{0.365}\ci{\mathbf{0.028}}$ \\
\midrule
\multirow{4}{*}{7VNX} & RMSD (\AA, $\downarrow$) & $-$ & $2.787\ci{1.086}$ & $\mathbf{0.635}\ci{\mathbf{0.014}}$ & $1.113\ci{0.040}$ & $\mathbf{0.590}\ci{\mathbf{0.060}}$ \\
 & C$\alpha$ RMSD (\AA, $\downarrow$) & $-$ & $2.387\ci{1.125}$ & $\mathbf{0.359}\ci{\mathbf{0.021}}$ & $\mathbf{0.290}\ci{\mathbf{0.098}}$ & $\mathbf{0.354}\ci{\mathbf{0.059}}$ \\
 & $R_\mathrm{work}$ ($\downarrow$) & 0.174 & $0.544\ci{0.003}$ & $0.552\ci{0.001}$ & $\mathbf{0.317}\ci{\mathbf{0.004}}$ & $\mathbf{0.328}\ci{\mathbf{0.016}}$ \\
 & $R_\mathrm{free}$ ($\downarrow$) & 0.207 & $0.569\ci{0.012}$ & $0.557\ci{0.005}$ & $\mathbf{0.321}\ci{\mathbf{0.008}}$ & $\mathbf{0.328}\ci{\mathbf{0.012}}$ \\
\midrule
\multirow{4}{*}{1L63} & RMSD (\AA, $\downarrow$) & $-$ & $1.784\ci{0.132}$ & $\mathbf{0.650}\ci{\mathbf{0.006}}$ & $0.940\ci{0.013}$ & $\mathbf{0.661}\ci{\mathbf{0.024}}$ \\
 & C$\alpha$ RMSD (\AA, $\downarrow$) & $-$ & $1.271\ci{0.084}$ & $\mathbf{0.201}\ci{\mathbf{0.031}}$ & $\mathbf{0.237}\ci{\mathbf{0.014}}$ & $\mathbf{0.220}\ci{\mathbf{0.067}}$ \\
 & $R_\mathrm{work}$ ($\downarrow$) & 0.148 & $0.535\ci{0.002}$ & $0.550\ci{0.005}$ & $\mathbf{0.327}\ci{\mathbf{0.003}}$ & $\mathbf{0.310}\ci{\mathbf{0.016}}$ \\
 & $R_\mathrm{free}$ ($\downarrow$) & 0.174 & $0.561\ci{0.011}$ & $0.586\ci{0.011}$ & $0.344\ci{0.003}$ & $\mathbf{0.309}\ci{\mathbf{0.019}}$ \\
\bottomrule
\end{tabular}%
}
\label{tab:exp_results}
\end{table}

\paragraph{Discussion.}
CrystalBoltz attains the lowest RMSD and R-factors on most of the cases and consistently improves the R-factors from the unguided predictions (Table~\ref{tab:exp_results}). The improvement is largest on the two targets with substantial conformational difference from to the prior: on 8DWN ($2.65$~\AA{} RMSD) and 4NTZ ($4.54$~\AA{} RMSD). Guiding conformational changes on such a large scale is extremely challenging due to the rugged optimization landscape. Nevertheless, CrystalBoltz reaches $1.32$~\AA{} and $1.30$~\AA{} RMSD against the strongest baseline values of $2.20$~\AA{} and $4.54$~\AA{}, with $R_\mathrm{free}$ reduced by $0.045$ and $0.062$. Visualization of 4NTZ (Fig.~\ref{fig:4ntz_qual_fig}) confirms that CrystalBoltz recovers the conformational rearrangement supported by the experimental data, whereas the baselines remain trapped near the initial generative model prediction, which accounts for their high RMSD and R-factors on this target. 
We also tested on targets (7O51, 7SEZ, 7VNX, 1L63) where the Boltz-2 priors are already close to the deposited structures. CrystalBoltz still reliably refines the predictions with experimental guidance in this regime, with smaller gains on RMSD as expected, but consistent and significant improvement on $R_\text{free}$.

Furthermore, on a single A6000 (see Table \ref{tab:runtime_head_to_head}), CrystalBoltz is $33.3\times$ faster than ROCKET, reducing a multi-hour job to minutes. This gap reflects the cost of each pipeline: ROCKET requires three phase-1 runs (each 100 iterations) of iterative MSA-bias optimization, a longer phase-2 (500 iterations) fine-tune, and a final \texttt{phenix.refine} step, while CrystalBoltz applies experimental guidance during reverse diffusion and a short Adam-based refinement (usually only 50 steps) against the crystallographic forward model. At this scale, experimental guidance becomes feasible during a beamtime session rather than as an offline post-processing step, opening the door to data-driven structure refinement in the same iteration loop as data collection. The combination of accuracy on hard conformational changes and minute-scale runtime is what makes guided diffusion competitive with classical refinement pipelines. More results and details on the implementation of \Method can be found in Appendix~\ref{more_results} and~\ref{hyperparam}.

\begin{table}[h!]
\centering
\caption{\textbf{Average runtime} on NVIDIA GPU RTX A6000 over 6 example proteins. For both methods, we use the authors' recommended number of refinement stages as specified in their respective implementation details. These runtimes correspond directly to the performance metrics reported in Table \ref{tab:exp_results}. \Method achieves a 33.3$\times$ speedup, reducing a multi-hour process to minutes.}
\vspace{0.5em}
\begin{tabular}{lrrrr}
\toprule
\multirow{2}{*}{\textbf{Method}} & \multicolumn{3}{c}{\textbf{Runtime (min)}} & \multirow{2}{*}{\textbf{Speedup}} \\
\cmidrule(lr){2-4}
 & Phase 1 & Phase 2 & \textbf{Total} & \\
\midrule
Rocket (Baseline) & 92.0 & 284.0 & 376.0 & 1$\times$ \\
\textbf{\Method (Ours)} & \textbf{10.9} & \textbf{0.4} & \textbf{11.3} & \textbf{33.3$\times$} \\
\bottomrule
\end{tabular}
\label{tab:runtime_head_to_head}
\end{table}


\subsection{Ablation Studies}
We ablate three components of \Method to isolate their individual contributions: the bulk-solvent term in the forward model, joint B-factor refinement, and the scheduling of guidance and post-hoc refinement.

\paragraph{Effect of solvent in the forward model.}
We ablate the bulk-solvent contribution in our differentiable forward
model by replacing $\mathbf{F}_\text{total}=\mathbf{F}_\text{protein}+\mathbf{F}_\text{solvent}$
with $\mathbf{F}_\text{protein}$ alone, and re-evaluate the Pearson correlation
against the experimental amplitudes $\mathbf{F}^{\,\text{exp}}_{\,\text{total}}$. Including the bulk-solvent term
yields a consistent improvement of $+0.090$, $+0.056$, and $+0.107$ in
CC for the unguided Boltz-2 baseline, our guided model (Ours), and the
deposited ground-truth structure (GT, included as a reference upper
bound), respectively. The improvement at GT is particularly
informative: with protein coordinates held fixed and correct, the
$+0.107$ gain isolates the contribution of the bulk-solvent term in
\texttt{SFCalculator} and confirms that the forward model is recovering
a non-trivial component of the experimental signal that a
protein-only model structurally cannot. 

\begin{figure}[H]
  \centering
  
  \begin{minipage}[b]{0.48\linewidth}
    \centering
    \includegraphics[width=\linewidth]{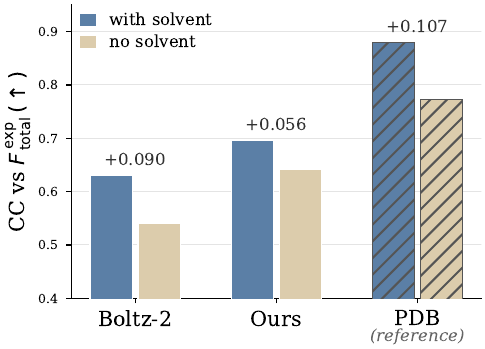}
    \caption{\textbf{Solvent ablation on PDB 8DWN.} Bulk-solvent term consistently improves CC.}
    \label{fig:solvent_ablate}
  \end{minipage}
  \hfill 
  \begin{minipage}[b]{0.48\linewidth}
    \centering
    \includegraphics[width=0.9\linewidth]{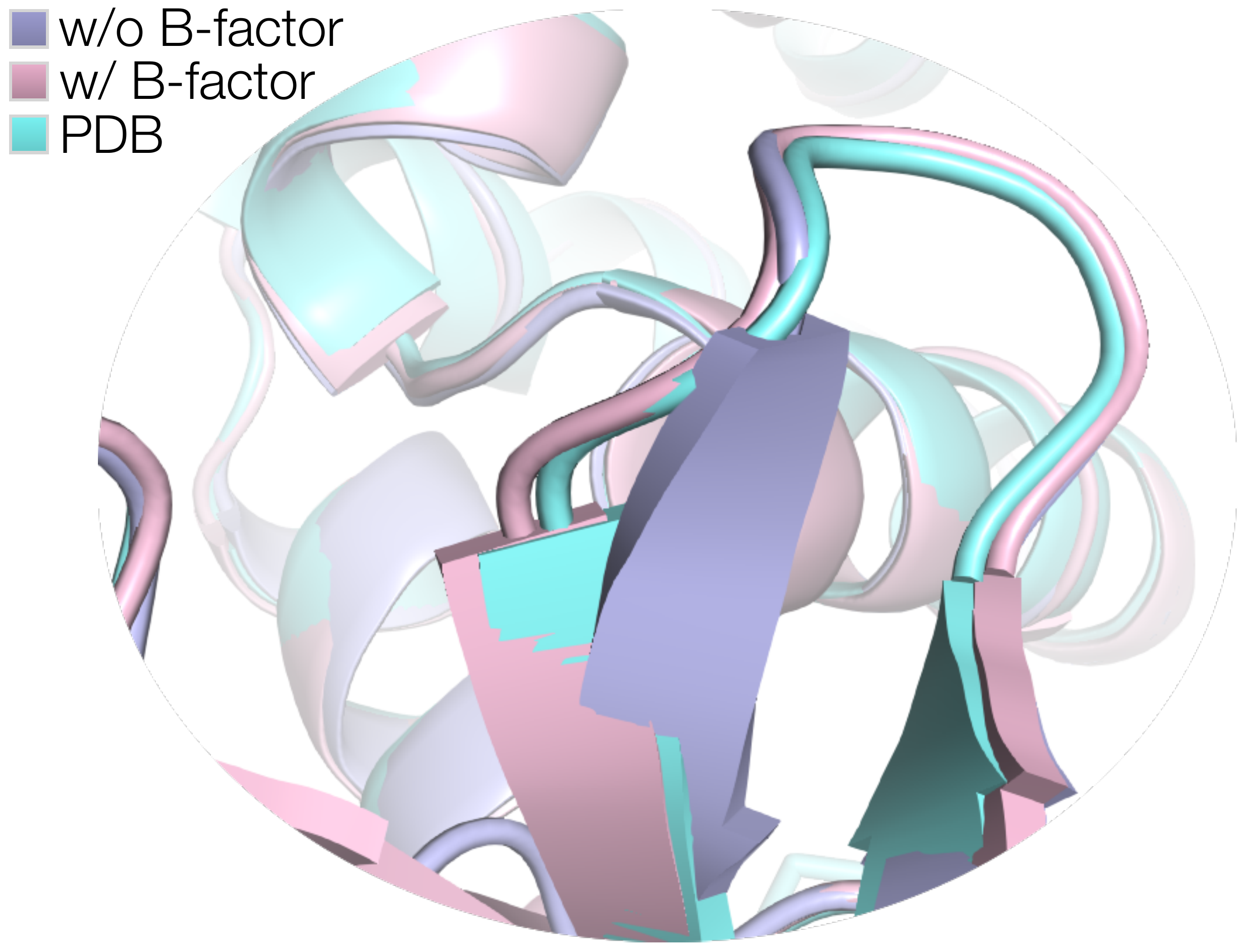}
    \caption{\textbf{Zoomed-in view of PDB 1L63.} B-factor optimization improves RMSD.}
    \label{fig:bfactor_zoom}
  \end{minipage}
  
\end{figure}

\begin{wraptable}{r}{0.4\textwidth}
  \centering
  \caption{\textbf{Effect of B-factor refinement} during the post-guidance refinement stage of \Method on PDB 1L63.}
  \label{tab:bfactor_ablation}
  \vspace{0.1em}
  \small
  \setlength{\tabcolsep}{4pt}
  \begin{tabular}{lcc}
    \toprule
    & w/o B-factor & w/ B-factor  \\
    \midrule
    RMSD (\AA) & 0.6726 & \textbf{0.6519}  \\
    C$\alpha$ RMSD (\AA) & 0.2434 & \textbf{0.2215}  \\
    $R_\text{work}$ & 0.3427 & 0.3434  \\
    $R_\text{free}$ & 0.3322 & 0.3323  \\
    \bottomrule
  \end{tabular}
\end{wraptable}

\paragraph{Effect of B-factor refinement.}
The post-guidance refinement stage of \Method jointly optimizes
atomic coordinates and isotropic B-factors against the experimental
amplitudes. Disabling B-factor refinement on 1L63, with all other
settings fixed, yields a small but consistent increase in both global
($+0.021$\,\AA) and C$\alpha$ ($+0.022$\,\AA) RMSD against the
deposited structure, while $R_\text{work}$ and $R_\text{free}$ are
effectively unchanged. With B-factors fixed, the only remaining degree of freedom for fitting
the amplitudes is atomic position, so the optimizer can only fit the
data by moving atoms, even when the real mismatch comes from atomic
motion that B-factors would otherwise capture. Refining B-factors absorbs
that disorder where it physically belongs, leaving the coordinates
free to relax toward the reference. The amplitude fit is comparable
in both settings; what changes is how the data signal is partitioned
between coordinates and B-factors.

\begin{table}[h]
\centering
\caption{\textbf{Ablation on guidance scheduling and post-hoc refinement with PDB 4NTZ.}
Our full pipeline (mid-way guidance with refinement) achieves the best score on every metric,
showing that delayed guidance and post-hoc refinement provide complementary, non-redundant gains.}
\label{tab:ablation}
\begin{tabular}{lcccc}
\toprule
Configuration & RMSD ($\downarrow$) & C$\alpha$ RMSD ($\downarrow$) & $R_{\text{work}}$ ($\downarrow$) & $R_{\text{free}}$ ($\downarrow$) \\
\midrule
Mid-way guidance + refinement (ours) & \textbf{1.2986} & \textbf{0.7687} & \textbf{0.4472} & \textbf{0.4825} \\
No guidance, refinement only         & 6.9013 & 6.4516 & 0.5215 & 0.5610 \\
Mid-way guidance, no refinement      & 2.8058 & 2.3711 & 0.5013 & 0.5518 \\
Full guidance + refinement           & 8.4199 & 7.9865 & 0.5399 & 0.5537 \\
\bottomrule
\end{tabular}
\end{table}

\paragraph{Effect of guidance and refinement parametrization.}
Table~\ref{tab:ablation} disentangles the two components of our pipeline. Applying guidance from the first denoising step (row 4) actually hurts performance relative to mid-way guidance: when the coordinates are still essentially noise, the structure-factor likelihood gradient is dominated by spurious contributions and pulls the trajectory toward poor minima before any backbone topology has formed. Delaying guidance until a rough fold has emerged (row 1) avoids this regime, so the experimental signal corrects an already-plausible structure rather than driving formation from scratch. Such a "warm-up" stage has also been found beneficial in cryo-EM guided structure prediction~\citep{raghu2025cryoboltz}. Post-guidance refinement addresses a complementary failure mode. As discussed in Sec. \ref{post-hoc_refine}, DPS guidance is constrained to a likelihood-based objective propagated through the denoiser Jacobian and operates only at the resolution of the generative prior, so it captures the global fold but cannot resolve local details. The refinement stage drops these constraints: it acts directly on $\hat{\mathbf{X}}_0$ via the crystallographic forward model alone and optimizes the evaluation metric itself, while the jointly fit B-factors absorb local disorder through the Debye--Waller envelope. Adding refinement on top of mid-way guidance (row 3 to row 1) lowers $R_{\text{work}}$ from $0.5013$ to $0.4472$ and $R_{\text{free}}$ from $0.5518$ to $0.4825$, confirming that this local fit is what closes the gap on the crystallographic metrics. The no-guidance baseline (row 2) shows the converse: refinement is a local optimizer with a narrow radius of convergence, and without a good initialization from guidance it cannot recover either the fold or the R-factors. Guidance and refinement therefore operate on different scales --- guidance establishes the correct fold, refinement performs the precise fit against the data --- and both are required.

\section{Conclusion}
We presented CrystalBoltz, an end-to-end framework that integrates experimental X-ray crystallographic data into protein structure determination by treating it as posterior sampling under a sequence-conditioned diffusion prior. The pipeline has two stages: experiment-guided diffusion driven by gradients from a differentiable forward model on structure-factor amplitudes, switched on partway through sampling once a coarse fold has emerged; and a short post-guidance refinement that jointly optimizes coordinates and B-factors against the crystallographic objective directly. On six experimental targets spanning a range of resolutions, sizes, and conformational distances from the prior, CrystalBoltz consistently improves the agreement between the predicted structures and the experimental data, attaining the lowest RMSD and R-factors on the majority of metric--target pairs, while running $33.3\times$ faster than ROCKET on identical hardware.

Several directions follow naturally from this work. A systematic evaluation on multi-chain and multi-copy assemblies is the most immediate next step, since the forward model and guidance loss are already chain-agnostic and Boltz-2 supports multi-chain prediction natively. On the algorithmic side, our pipeline uses DPS~\cite{dps} as the posterior sampling method, but recent diffusion-based inverse-problem methods~\cite{zhang2024daps,ddiff2026} have shown stronger results on noisy and nonlinear inverse problems; substituting them in our framework is a direct test of whether the remaining gap to the deposited PDB R-factors is limited by guidance algorithm. Furthermore, the same posterior-sampling formulation readily extends to other experimental modalities with differentiable forward models, such as real-space refinement against cryo-EM density maps. We recognize that the dependency on reference structures for alignment is a common limitation in experimental guidance, including our work. Future research on a more comprehensive and robust forward modeling process could address this limitation.


\section*{Acknowledgements}
MK, AF and FP were supported by the Department of Energy, Laboratory Directed Research and Development program at SLAC National Accelerator Laboratory, under contract DE-AC02-76SF00515. MK, GW and FP acknowledge support from the Stanford HAI Seed Grant Program.

\newpage
\bibliographystyle{plainnat}
\bibliography{neurips_2026}

@misc{survey_posterior_sampling,
      title={A Survey on Diffusion Models for Inverse Problems},
      author={Giannis Daras and Hyungjin Chung and Chieh-Hsin Lai and Yuki Mitsufuji and Jong Chul Ye and Peyman Milanfar and Alexandros G. Dimakis and Mauricio Delbracio},
      year={2024},
      eprint={2410.00083},
      archivePrefix={arXiv},
      primaryClass={cs.LG},
      url={https://arxiv.org/abs/2410.00083},
}

@article{sfcalculator,
  title={SFCalculator: connecting deep generative models and crystallography},
  author={Li, Minhuan and Dalton, Kevin M and Hekstra, Doeke Romke},
  journal={bioRxiv},
  pages={2025--01},
  year={2025},
  publisher={Cold Spring Harbor Laboratory}
}

@inproceedings{
maddipatla2025inverse,
title={Inverse problems with experiment-guided AlphaFold},
author={Sai Advaith Maddipatla and Nadav Bojan and Meital Bojan and Sanketh Vedula and Ailie Marx and Paul Schanda and Alexander Bronstein},
booktitle={ICLR 2025 Workshop on Generative and Experimental Perspectives for Biomolecular Design},
year={2025},
url={https://openreview.net/forum?id=1gp130uxfw}
}

@article{read2016log,
  title={A log-likelihood-gain intensity target for crystallographic phasing that accounts for experimental error},
  author={Read, Randy J and McCoy, Airlie J},
  journal={Biological Crystallography},
  volume={72},
  number={3},
  pages={375--387},
  year={2016},
  publisher={International Union of Crystallography}
}

@article{winter2010xia2,
  title={xia2: an expert system for macromolecular crystallography data reduction},
  author={Winter, Graeme},
  journal={Journal of Applied Crystallography},
  volume={43},
  number={1},
  pages={186--190},
  year={2010},
  doi={10.1107/S0021889809045701},
  publisher={International Union of Crystallography}
}

@article{hendrickson2023phaseproblem,
  title={Facing the phase problem},
  author={Hendrickson, Wayne A},
  journal={IUCrJ},
  volume={10},
  number={5},
  pages={521--543},
  year={2023},
  doi={10.1107/S2052252523006449},
  publisher={International Union of Crystallography}
}

@article{taylor2003phase,
  title={The phase problem},
  author={Taylor, Garry},
  journal={Biological Crystallography},
  volume={59},
  number={11},
  pages={1881--1890},
  year={2003},
  publisher={International Union of Crystallography}
}

@article{afonine2013bulk,
  title={Bulk-solvent and overall scaling revisited: faster calculations, improved results},
  author={Afonine, PV and Grosse-Kunstleve, RW and Adams, PD and Urzhumtsev, A},
  journal={Biological Crystallography},
  volume={69},
  number={4},
  pages={625--634},
  year={2013},
  publisher={International Union of Crystallography}
}

@inproceedings{
dps,
title={Diffusion Posterior Sampling for General Noisy Inverse Problems},
author={Hyungjin Chung and Jeongsol Kim and Michael Thompson Mccann and Marc Louis Klasky and Jong Chul Ye},
booktitle={The Eleventh International Conference on Learning Representations },
year={2023},
url={https://openreview.net/forum?id=OnD9zGAGT0k}
}

@article{
baek2021,
author = {Minkyung Baek  and Frank DiMaio  and Ivan Anishchenko  and Justas Dauparas  and Sergey Ovchinnikov  and Gyu Rie Lee  and Jue Wang  and Qian Cong  and Lisa N. Kinch  and R. Dustin Schaeffer  and Claudia Millán  and Hahnbeom Park  and Carson Adams  and Caleb R. Glassman  and Andy DeGiovanni  and Jose H. Pereira  and Andria V. Rodrigues  and Alberdina A. van Dijk  and Ana C. Ebrecht  and Diederik J. Opperman  and Theo Sagmeister  and Christoph Buhlheller  and Tea Pavkov-Keller  and Manoj K. Rathinaswamy  and Udit Dalwadi  and Calvin K. Yip  and John E. Burke  and K. Christopher Garcia  and Nick V. Grishin  and Paul D. Adams  and Randy J. Read  and David Baker },
title = {Accurate prediction of protein structures and interactions using a three-track neural network},
journal = {Science},
volume = {373},
number = {6557},
pages = {871-876},
year = {2021},
doi = {10.1126/science.abj8754},
URL = {https://www.science.org/doi/abs/10.1126/science.abj8754},
eprint = {https://www.science.org/doi/pdf/10.1126/science.abj8754}}

@article{pdb,
  author    = {Berman, Helen M. and Westbrook, John and Feng, Zukang and Gilliland, Gary and Bhat, T. N. and Weissig, Helge and Shindyalov, Ilya N. and Bourne, Philip E.},
  title     = {The {Protein} {Data} {Bank}},
  journal   = {Nucleic Acids Research},
  volume    = {28},
  number    = {1},
  pages     = {235--242},
  year      = {2000},
  doi       = {10.1093/nar/28.1.235},
}

@article{jumper2021,
  author = {Jumper, John and Evans, Richard and Pritzel, Alexander and Green, Tim and Figurnov, Michael and Ronneberger, Olaf and Tunyasuvunakool, Kathryn and Bates, Russ and Z{\i}dek, Augustin and Potapenko, Anna and Bridgland, Alex and Meyer, Clemens and Kohl, Simon A. A. and Ballard, Andrew J. and Cowie, Andrew and Romera-Paredes, Bernardino and Nikolov, Stanislav and Jain, Rishub and Adler, Jonas and Back, Trevor and Petersen, Stig and Reiman, David and Clancy, Ellen and Zielinski, Michal and Steinegger, Martin and Pacholska, Michalina and Berghammer, Tamas and Bodenstein, Sebastian and Silver, David and Vinyals, Oriol and Senior, Andrew W. and Kavukcuoglu, Koray and Kohli, Pushmeet and Hassabis, Demis},
  title = {Highly accurate protein structure prediction with AlphaFold},
  journal = {Nature},
  volume = {596},
  number = {7873},
  pages = {583--589},
  year = {2021},
  doi = {10.1038/s41586-021-03819-2}
}

@article{alphafold3,
  author = {Abramson, Josh and Adler, Jonas and Dunger, Jack and Evans, Richard and Green, Tim and Pritzel, Alexander and Ronneberger, Olaf and Willmore, Laurence and Ballard, Andrew J. and Bambrick, Joseph and Bodenstein, Sebastian and others},
  title = {Accurate structure prediction of biomolecular interactions with AlphaFold 3},
  journal = {Nature},
  year = {2024},
  doi = {10.1038/s41586-024-07487-w}
}

@article{terwilliger2023,
  author = {Terwilliger, Thomas C. and others},
  title = {AlphaFold predictions are valuable hypotheses and accelerate but do not replace experimental structure determination},
  journal = {Nature Methods},
  year = {2023},
  doi = {10.1038/s41592-023-02031-9}
}

@article{murshudov1997,
  author = {Murshudov, G. N. and Vagin, A. A. and Dodson, E. J.},
  title = {Refinement of macromolecular structures by the maximum-likelihood method},
  journal = {Acta Crystallographica Section D: Biological Crystallography},
  volume = {53},
  number = {3},
  pages = {240--255},
  year = {1997},
  doi = {10.1107/S0907444996012255}
}

@article{refmac5,
  author = {Murshudov, G. N. and Skub{\'a}k, P. and Lebedev, A. A. and Pannu, N. S. and Steiner, R. A. and Nicholls, R. A. and Winn, M. D. and Long, F. and Vagin, A. A.},
  title = {REFMAC5 for the refinement of macromolecular crystal structures},
  journal = {Acta Crystallographica Section D: Biological Crystallography},
  volume = {67},
  number = {4},
  pages = {355--367},
  year = {2011},
  doi = {10.1107/S0907444911001314}
}

@article{phenix2019,
  author = {Liebschner, Dorothee and Afonine, Pavel V. and Baker, Matthew L. and Bunk{\'o}czi, G{\'a}bor and Chen, Vincent B. and Croll, Tristan I. and Hintze, Bradley and Hung, Li-Wei and Jain, Swati and McCoy, Airlie J. and Moriarty, Nigel W. and Oeffner, Robert D. and Poon, Billy K. and Prisant, Mikhail G. and Read, Randy J. and Richardson, Jane S. and Richardson, David C. and Sammito, Michael D. and Sobolev, Oleg V. and Stockwell, Daniel H. and Terwilliger, Thomas C. and Urzhumtsev, Alexandre and Videau, Lucie L. and Williams, Christopher J. and Adams, Paul D.},
  title = {Macromolecular structure determination using X-rays, neutrons and electrons: recent developments in Phenix},
  journal = {Acta Crystallographica Section D: Structural Biology},
  volume = {75},
  number = {10},
  pages = {861--877},
  year = {2019},
  doi = {10.1107/S2059798319011471}
}

@article{fadini2025rocket,
  author = {Fadini, Alisia and Li, Minhuan and McCoy, Airlie J. and Terwilliger, Thomas C. and Read, Randy J. and Hekstra, Doeke R. and AlQuraishi, Mohammed},
  title = {AlphaFold as a prior: experimental structure determination conditioned on a pretrained neural network},
  journal = {Nature Methods},
  volume = {23},
  number = {7},
  pages = {785--795},
  year = {2026},
  doi = {10.1038/s41592-026-03047-4}
}

@article{brunger1992free,
  author  = {Br{\"u}nger, Axel T},
  title   = {Free {R} value: a novel statistical quantity for assessing
             the accuracy of crystal structures},
  journal = {Nature},
  volume  = {355},
  number  = {6359},
  pages   = {472--475},
  year    = {1992},
}

@article{boltz1,
  author  = {Wohlwend, Jeremy and Corso, Gabriele and Passaro, Saro and Reveiz, Mateo and Leidal, Ken and Swiderski, Wojtek and Portnoi, Tally and Chinn, Itamar and Silterra, Jacob and Jaakkola, Tommi and Barzilay, Regina},
  title   = {Boltz-1: Democratizing Biomolecular Interaction Modeling},
  journal = {bioRxiv},
  year    = {2024},
  doi     = {10.1101/2024.11.19.624167},
  url     = {https://www.biorxiv.org/content/10.1101/2024.11.19.624167v2}
}

@article{boltz2,
  author  = {Passaro, Saro and Corso, Gabriele and Wohlwend, Jeremy and
             Reveiz, Mateo and Thaler, Stephan and Somnath, Vignesh Ram
             and Getz, Noah and Portnoi, Tally and Roy, Julien and
             Stark, Hannes and Kwabi-Addo, David and Beaini, Dominique
             and Jaakkola, Tommi and Barzilay, Regina},
  title   = {Boltz-2: Towards Accurate and Efficient Binding Affinity
             Prediction},
  journal = {bioRxiv},
  year    = {2025},
}

@inproceedings{zhang2024daps,
  title={Improving Diffusion Inverse Problem Solving with Decoupled Noise Annealing},
  author={Zhang, Bingliang and Chu, Wenda and Berner, Julius and Meng, Chenlin and Anandkumar, Anima and Song, Yang},
  booktitle={International Conference on Learning Representations},
  year={2025}
}

@inproceedings{ddiff2026,
  title={Dual Ascent Diffusion for Inverse Problems},
  author={Kim, Minseo and Levy, Axel and Wetzstein, Gordon},
  booktitle={IEEE/CVF Conference on Computer Vision and Pattern Recognition},
  year={2026}
}

@article{french1978wilson,
  author  = {French, S. and Wilson, K.},
  title   = {On the Treatment of Negative Intensity Observations},
  journal = {Acta Crystallographica Section A},
  volume  = {34},
  number  = {4},
  pages   = {517--525},
  year    = {1978},
  doi     = {10.1107/S0567739478001114}
}

@article{read1990rice,
  author  = {Read, Randy J.},
  title   = {Structure-Factor Probabilities for Related Structures},
  journal = {Acta Crystallographica Section A},
  volume  = {46},
  number  = {11},
  pages   = {900--912},
  year    = {1990},
  doi     = {10.1107/S0108767390005529}
}

@book{rhodes2006crystallography,
  author    = {Rhodes, Gale},
  title     = {Crystallography Made Crystal Clear: A Guide for Users of Macromolecular Models},
  edition   = {3rd},
  publisher = {Academic Press},
  series    = {Complementary Science},
  address   = {Amsterdam},
  year      = {2006},
  isbn      = {978-0-12-587073-3}
}

@book{rupp2009biomolecular,
  author    = {Rupp, Bernhard},
  title     = {Biomolecular Crystallography: Principles, Practice, and Application to Structural Biology},
  edition   = {1st},
  publisher = {Garland Science},
  address   = {New York},
  year      = {2009},
  isbn      = {9780429258756}
}

@inproceedings{raghu2025cryoboltz,
        title       = {Multiscale guidance of protein structure prediction with heterogeneous cryo-EM data},
        author      = {Rishwanth Raghu and Axel Levy and Gordon Wetzstein and Ellen D. Zhong},
	booktitle   = {NeurIPS},
        year        = {2025}
      }

@article{keegan2024success,
  title={The success rate of processed predicted models in molecular replacement: implications for experimental phasing in the AlphaFold era},
  author={Keegan, Ronan M and Simpkin, Adam J and Rigden, Daniel J},
  journal={Biological Crystallography},
  volume={80},
  number={11},
  year={2024},
  publisher={International Union of Crystallography}
}

@article{oeffner2022putting,
  title={Putting AlphaFold models to work with phenix. process\_predicted\_model and ISOLDE},
  author={Oeffner, Robert D and Croll, Tristan I and Mill{\'a}n, Claudia and Poon, Billy K and Schlicksup, Christopher J and Read, Randy J and Terwilliger, Tom C},
  journal={Biological Crystallography},
  volume={78},
  number={11},
  pages={1303--1314},
  year={2022},
  publisher={International Union of Crystallography}
}

@article{rf3,
  title={De novo design of all-atom biomolecular interactions with rfdiffusion3},
  author={Butcher, Jasper and Krishna, Rohith and Mitra, Raktim and Brent, Rafael I and Li, Yanjing and Corley, Nathaniel and Kim, Paul T and Funk, Jonathan and Mathis, Simon and Salike, Saman and others},
  journal={bioRxiv},
  year={2025}
}

@article{protenix,
  title={Protenix-advancing structure prediction through a comprehensive AlphaFold3 reproduction},
  author={ByteDance AML AI4Science Team and Chen, Xinshi and Zhang, Yuxuan and Lu, Chan and Ma, Wenzhi and Guan, Jiaqi and Gong, Chengyue and Yang, Jincai and Zhang, Hanyu and Zhang, Ke and others},
  journal={BioRxiv},
  pages={2025--01},
  year={2025},
  publisher={Cold Spring Harbor Laboratory}
}

@book{srinivasan1976some,
  title     = {Some Statistical Applications in X-ray Crystallography},
  author    = {Srinivasan, Ramachandran and Parthasarathy, Soundarajan},
  year      = {1976},
  publisher = {Pergamon Press},
  address   = {Oxford},
  isbn      = {0080180469},
}

@article{ahdritz2024openfold,
  title={OpenFold: retraining AlphaFold2 yields new insights into its learning mechanisms and capacity for generalization},
  author={Ahdritz, Gustaf and Bouatta, Nazim and Floristean, Christina and Kadyan, Sachin and Xia, Qinghui and Gerecke, William and O’Donnell, Timothy J and Berenberg, Daniel and Fisk, Ian and Zanichelli, Niccol{\`o} and others},
  journal={Nature methods},
  volume={21},
  number={8},
  pages={1514--1524},
  year={2024},
  publisher={Nature Publishing Group US New York}
}

@article{afonine2012towards,
  title={Towards automated crystallographic structure refinement with phenix. refine},
  author={Afonine, Pavel V and Grosse-Kunstleve, Ralf W and Echols, Nathaniel and Headd, Jeffrey J and Moriarty, Nigel W and Mustyakimov, Marat and Terwilliger, Thomas C and Urzhumtsev, Alexandre and Zwart, Peter H and Adams, Paul D},
  journal={Biological crystallography},
  volume={68},
  number={4},
  pages={352--367},
  year={2012},
  publisher={International Union of Crystallography}
}

@article{winter2018dials,
  title={DIALS: implementation and evaluation of a new integration package},
  author={Winter, Graeme and Waterman, David G and Parkhurst, James M and Brewster, Aaron S and Gildea, Richard J and Gerstel, Markus and Fuentes-Montero, Luis and Vollmar, Melanie and Michels-Clark, Tara and Young, Iris D and others},
  journal={Biological Crystallography},
  volume={74},
  number={2},
  pages={85--97},
  year={2018},
  publisher={International Union of Crystallography}
}

@article{kabsch2010xds,
  title={xds},
  author={Kabsch, Wolfgang},
  journal={Biological crystallography},
  volume={66},
  number={2},
  pages={125--132},
  year={2010},
  publisher={International Union of Crystallography}
}

@article{wayment2024predicting,
  title={Predicting multiple conformations via sequence clustering and AlphaFold2},
  author={Wayment-Steele, Hannah K and Ojoawo, Adedolapo and Otten, Renee and Apitz, Julia M and Pitsawong, Warintra and H{\"o}mberger, Marc and Ovchinnikov, Sergey and Colwell, Lucy and Kern, Dorothee},
  journal={Nature},
  volume={625},
  number={7996},
  pages={832--839},
  year={2024},
  publisher={Nature Publishing Group UK London}
}

@article{kalakoti2025afsample2,
  title={AFsample2 predicts multiple conformations and ensembles with AlphaFold2},
  author={Kalakoti, Yogesh and Wallner, Bj{\"o}rn},
  journal={Communications biology},
  volume={8},
  number={1},
  pages={373},
  year={2025},
  publisher={Nature Publishing Group UK London}
}

@article{del2022sampling,
  title={Sampling alternative conformational states of transporters and receptors with AlphaFold2},
  author={Del Alamo, Diego and Sala, Davide and Mchaourab, Hassane S and Meiler, Jens},
  journal={elife},
  volume={11},
  pages={e75751},
  year={2022},
  publisher={eLife Sciences Publications, Ltd}
}

@article{stein2022speach_af,
  title={SPEACH\_AF: Sampling protein ensembles and conformational heterogeneity with Alphafold2},
  author={Stein, Richard A and Mchaourab, Hassane S},
  journal={PLoS computational biology},
  volume={18},
  number={8},
  pages={e1010483},
  year={2022},
  publisher={Public Library of Science San Francisco, CA USA}
}

@article{mccoy2022alphafold_phasing,
  title={Implications of AlphaFold2 for crystallographic phasing by molecular replacement},
  author={McCoy, Airlie J and Sammito, Massimo D and Read, Randy J},
  journal={Acta Crystallographica Section D: Structural Biology},
  volume={78},
  number={1},
  pages={1--13},
  year={2022},
  doi={10.1107/S2059798321012122}
}

@article{wang2025alphafold_guided_mr,
  title={AlphaFold-guided molecular replacement for solving challenging crystal structures},
  author={Wang, Wei and Gong, Zhening and Hendrickson, Wayne A},
  journal={Acta Crystallographica Section D: Structural Biology},
  volume={81},
  pages={4--21},
  year={2025},
  doi={10.1107/S2059798324011999}
}

@misc{levy2024protein_space_inverse,
  title={Solving Inverse Problems in Protein Space Using Diffusion-Based Priors},
  author={Levy, Axel and Chan, Eric R. and Fridovich-Keil, Sara and Poitevin, Frederic and Zhong, Ellen D. and Wetzstein, Gordon},
  year={2024},
  eprint={2406.04239},
  archivePrefix={arXiv},
  primaryClass={cs.LG},
  url={https://arxiv.org/abs/2406.04239}
}

@article{li2026robust,
  title={Robust Inference-Time Steering of Protein Diffusion Models via Embedding Optimization},
  author={Li, Minhuan and Han, Jiequn and Cossio, Pilar and Wu, Luhuan},
  journal={arXiv preprint arXiv:2602.05285},
  year={2026}
}

\newpage
\appendix
\section{Technical Appendices and Supplementary Material}
\subsection{Rice distribution likelihood}
\label{rice_dist}

This section provides the derivation and definitions underlying the Rice likelihood used in Eq.~\ref{eq:rice_loss} of the main text. We work with normalized structure-factor amplitudes $|\mathbf{E}_o(\vec{h})|$ and $|\mathbf{E}_c(\vec{h})|$, obtained by dividing $|\mathbf{F}_o|$ and $|\mathbf{F}_c|$ by their expected magnitudes within each resolution shell so that $\mathbb{E}[|\mathbf{E}_o(\vec{h})|^2] = \mathbb{E}[|\mathbf{E}_c(\vec{h})|^2] = 1$. Working in $\mathbf{E}$-values absorbs the resolution-dependent falloff of structure-factor magnitudes and removes the need for an explicit per-shell scaling term in the likelihood.

\paragraph{Acentric vs.\ centric reflections.} A complex structure factor $\mathbf{F}(\vec{h}) = A(\vec{h}) + i B(\vec{h})$ is the sum of contributions from many atoms. By the central limit theorem, for a generic Miller index the real and imaginary parts $A(\vec{h})$ and $B(\vec{h})$ are approximately independent Gaussians with equal variance, so $\mathbf{F}(\vec{h})$ follows a complex normal distribution; these are the \emph{acentric} reflections. For Miller indices invariant under a reciprocal-space symmetry operation that maps $\vec{h} \to -\vec{h}$, the crystallographic symmetry restricts the phase to two values $\{\varphi_0, \varphi_0 + \pi\}$, where $\varphi_0$ is fixed by the space group and the Miller index; the structure factor is then confined to a single line in the complex plane and projects onto a one-dimensional real Gaussian. These are the \emph{centric} reflections. The marginalization below is symmetric in $\varphi_0$, so $\varphi_0$ does not appear explicitly in the final centric density and is never computed at runtime; only the binary centric/acentric classification is needed. We compute this mask once at the start of guidance by enumerating the symmetry operations of the space group and flagging reflections for which some rotation $R$ satisfies $R^\top \vec{h} = -\vec{h}$.

\paragraph{Marginalizing the unobserved phase.} The experiment measures only the amplitude $|\mathbf{E}_o(\vec{h})|$, while the model produces both an amplitude and a phase $|\mathbf{E}_c(\vec{h})| e^{i\varphi_c(\vec{h})}$. The likelihood of the observation is therefore obtained by integrating the joint complex-normal density over the unobserved experimental phase. For an acentric reflection, integrating a two-dimensional Gaussian over the angular coordinate yields the Rice density; for a centric reflection, the analogous marginalization over the two allowed phase values yields a folded normal density. The $\sigma_A$-parameterized form below absorbs atomic coordinate error into the variance and incorporates per-reflection experimental error~\citep{read1990rice, read2016log}, building on the underlying Wilson statistics for structure-factor amplitudes~\citep{french1978wilson}. Concretely:
\begin{align}
    p_a \big(|\mathbf{E}_o(\vec{h})|;\, |\mathbf{E}_c(\vec{h})|\big) &= \frac{2|\mathbf{E}_o(\vec{h})|}{\Sigma_{\vec{h}}^2} \exp \!\left[ -\frac{|\mathbf{E}_o(\vec{h})|^2 + \big(\sigma_A |\mathbf{E}_c(\vec{h})|\big)^2}{\Sigma_{\vec{h}}^2} \right] I_0 \!\left( \frac{2\sigma_A |\mathbf{E}_o(\vec{h})| |\mathbf{E}_c(\vec{h})|}{\Sigma_{\vec{h}}^2} \right), \\
    p_c \big(|\mathbf{E}_o(\vec{h})|;\, |\mathbf{E}_c(\vec{h})|\big) &= \left[ \frac{2}{\pi \Sigma_{\vec{h}}^2} \right]^{1/2} \exp \!\left[ -\frac{|\mathbf{E}_o(\vec{h})|^2 + \big(\sigma_A |\mathbf{E}_c(\vec{h})|\big)^2}{2 \Sigma_{\vec{h}}^2} \right] \cosh \!\left( \frac{\sigma_A |\mathbf{E}_o(\vec{h})| |\mathbf{E}_c(\vec{h})|}{\Sigma_{\vec{h}}^2} \right),
\end{align}
where $I_0$ is the modified Bessel function of the first kind of order zero, arising from the angular integral $\frac{1}{2\pi}\int_0^{2\pi} \exp(z\cos\theta)\,d\theta = I_0(z)$ in the acentric case, and $\cosh$ arises analogously from the two-point average $\tfrac{1}{2}(e^{z}+e^{-z})$ over the two allowed centric phases.

\paragraph{Variance terms.} The per-reflection variance combines two sources of uncertainty:
\begin{equation}
    \Sigma_{\vec{h}}^2 =
    \begin{cases}
        1 - \sigma_A^2 + 2\tilde{\sigma}_{\vec{h}}^2 & (\vec{h} \in A), \\
        1 - \sigma_A^2 + \tilde{\sigma}_{\vec{h}}^2  & (\vec{h} \in C).
    \end{cases}
\end{equation}
The term $1 - \sigma_A^2$ accounts for model error: $\sigma_A \in [0, 1]$ is a Luzzati-style correlation parameter for the agreement between calculated and true structure factors, with $\sigma_A = 1$ a perfect model and $\sigma_A = 0$ an uninformative one. The term $\tilde{\sigma}_{\vec{h}}^2$ is the experimental measurement variance reported in the MTZ file, rescaled by $\mathrm{RMS}(|\mathbf{F}_o|)$ so that it is on the same $\mathbf{E}$-value scale as $|\mathbf{E}_o(\vec{h})|$ and $|\mathbf{E}_c(\vec{h})|$. The factor of 2 in the acentric case reflects the two real degrees of freedom of a complex Gaussian, versus one for the real-valued centric case. In our experiments we fix $\sigma_A = 0.85$ globally, consistent with the typical model-error regime for AlphaFold-quality starting models~\citep{terwilliger2023}; refining $\sigma_A$ jointly with coordinates is straightforward but did not improve results in our setup.

\subsection{Rigid body alignment}
\label{rigid_body_supp}
A prerequisite for each likelihood evaluation is aligning the denoiser prediction $\hat{\mathbf{X}}_0$ to the crystal frame before it is passed to \texttt{SFCalculator}. Boltz-2 operates in an SE(3)-augmented frame that is randomly rotated and translated at every sampling step; without explicit realignment, the fractional coordinates would be in an arbitrary orientation relative to the crystal axes, causing \texttt{SFCalculator} to compute entirely erroneous structure factor amplitudes. We therefore solve for the optimal rigid transformation
\begin{equation}
    (R^*, \mathbf{t}^*) = \operatorname*{arg\,min}_{R \in \mathrm{SO}(3),\, \mathbf{t}} \left\| R\hat{\mathbf{X}}_0 + \mathbf{t} - \mathbf{X}_{\mathrm{ref}} \right\|^2_W,
\end{equation}
where $\mathbf{X}_{\mathrm{ref}}$ is a fixed reference structure in the correct crystal setting and $\|\cdot\|^2_W$ denotes a per-atom weighted norm. The aligned coordinates $R^*\hat{\mathbf{X}}_0 + \mathbf{t}^*$ are then converted to fractional form via the unit cell $\mathbf{u}$ before computing $\mathbf{F}_c$. In practice, $\mathbf{X}_{\mathrm{ref}}$ is obtained by performing molecular replacement on the experimental MTZ file to place a Boltz-2-predicted structure in the correct crystal setting. While this is sufficient for the structures studied here, the quality of \Method is inherently tied to the accuracy of the molecular replacement step, which remains an active area of research and a promising direction for future improvement.

\subsection{Additional results}
\label{more_results}
In this section, we provide additional qualitative results.

\begin{figure}[h]
\centering
\includegraphics[width=\linewidth]{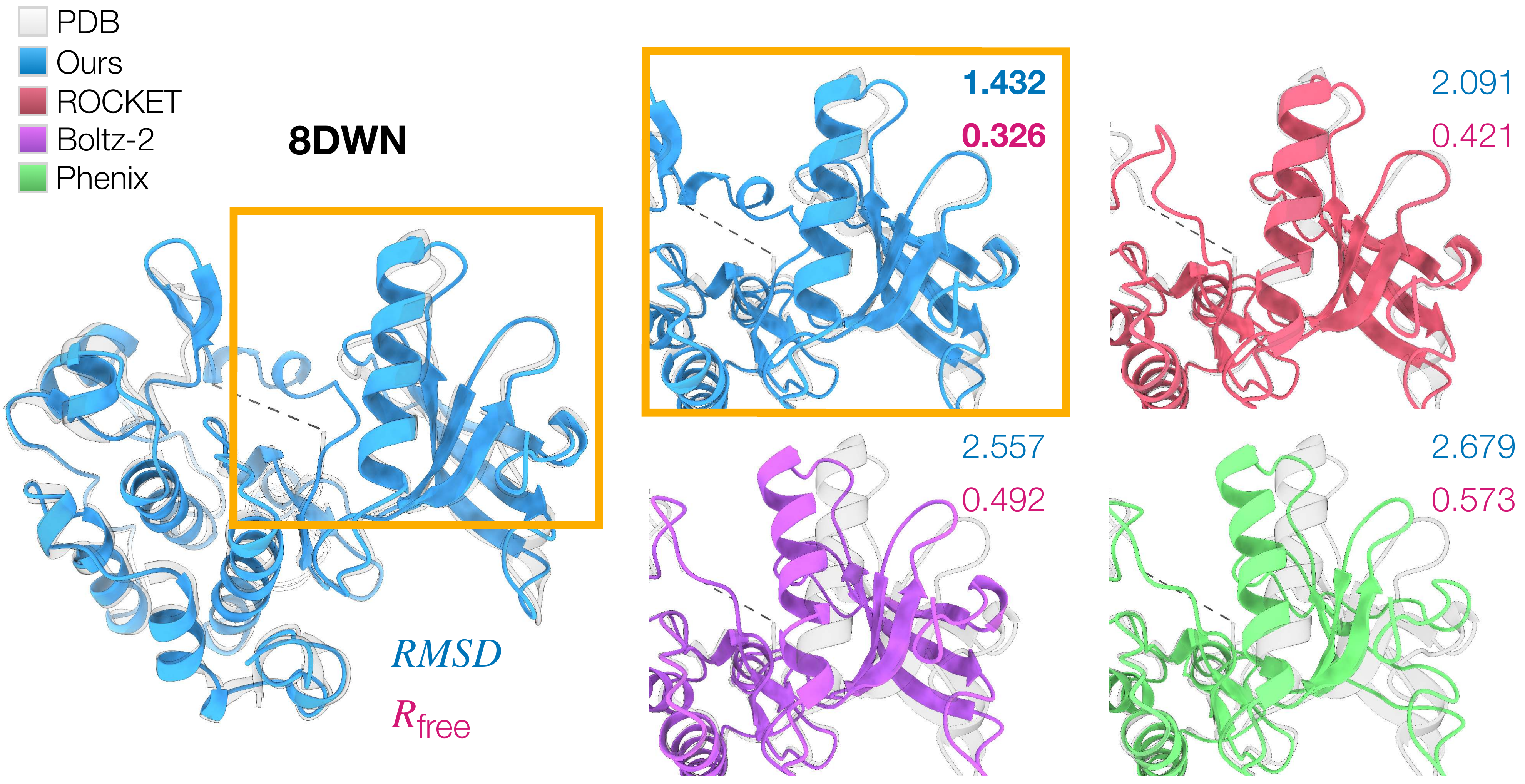}
\caption{\textbf{Qualitative results on PDB 8DWN.} Another example showing that \Method can correct large conformation change.}
\label{fig:8dwn_qual_fig}
\end{figure}

\subsection{Hyperparameter choices}
\label{hyperparam}
With the total number of diffusion sampling steps being $T=200$ for phase 1 of \Method, we chose the guidance start time as $t_g=50$ which is when the backbone structure is roughly recovered by the Boltz-2 prior. We recommend visual assessment of the structure at different time steps to choose this parameter but generally, including the protein targets we demonstrated in the main results (Table~\ref{tab:exp_results}), $t_g=50$ worked well.

The resolution-dependent scale factors $k_{\text{mask}}$ and $k_{\text{total}}$ enter the model amplitude as
\[
F_{\text{total}}(\mathbf{h}) \;=\; k_{\text{total}} \exp\!\big(-2\pi^{2}\mathbf{s}^{\top}\mathbf{U}_{\text{aniso}}\mathbf{s}\big)\big(F_{\text{protein}}(\mathbf{h}) + k_{\text{mask}}\,F_{\text{mask}}(\mathbf{h})\big).
\]
We discretize reciprocal space into ten log-spaced resolution bins. Both factors can in principle be fit per bin against $\mathbf{F}_{\text{o}}$: an analytical root-finding solver for $(k_{\text{total}}, k_{\text{mask}})$ following the cubic formulation of Afonine et al.~\cite{afonine2013bulk} is re-implemented in \texttt{SFCalculator}. During the post-DPS refinement stage we invoke this solver every 10 steps to track the evolving model. For the guidance stage itself, however, we found that a single constant initialization is enough: setting $k_{\text{total}}=0.5$ and $k_{\text{mask}}=0.35$ uniformly across all ten bins and across all six targets gave essentially indistinguishable $R$-factor and RMSD outcomes compared to running the per-bin solver at every guided step. The value $k_{\text{mask}}=0.35$ matches the canonical bulk-solvent contribution magnitude used by standard refinement packages such as \texttt{phenix.refine}. The value $k_{\text{total}}=0.5$ compensates for $\mathbf{F}_{\text{c}}$ from the Boltz-2 prior being systematically larger than $\mathbf{F}_{\text{o}}$ at the early guided steps, before the prior has relaxed onto the data manifold. The reason a constant scale is enough during guidance is that the likelihood gradient is dominated by the $|\mathbf{F}_{\text{o}}|-|\mathbf{F}_{\text{c}}|$ residuals, not the absolute scale; the per-bin Afonine fit only matters once the structure is clean enough for sub-percent $R$-factor differences to be meaningful, which is why we reserve it for refinement. Additionally, here $\mathbf{s}$ denotes the scattering vector associated with the Miller index $\mathbf{h}$, expressed in an orthonormal reciprocal-space basis. We initialize $\mathbf{U}_{\text{aniso}}$ as a scalar multiple of the identity matrix, which reduces the anisotropic envelope to an isotropic.

The guidance strength $\rho$ controls the per-step contribution of $ \nabla_{\mathbf{X}_t} \log p(\mathbf{y}|\mathbf{X}_t, \mathbf{a}, \mathbf{c})$ in Eq.~\ref{log_likeli} relative to the unconditional Boltz-2 score. The right value depends on the magnitude of $\nabla_{\mathbf{X}_t}\big(\lambda_{\text{gauss}} L_{\text{gauss}} + \lambda_{\text{rice}} L_{\text{rice}}\big)$ at $t_g$, which itself varies with the number of reflections, the working-set size, and how far the prior's $\mathbf{F}_{\text{c}}$ sits from $\mathbf{F}_{\text{o}}$. In practice we tune $\rho$ within $[10^{-3}, 10^{-1}]$ by inspecting the early guided trajectory: $\rho$ too large collapses geometry (bond-length and clash violations spike within a few steps), and $\rho$ too small leaves the final structure indistinguishable from the unconditional prior. We landed on $\rho = 0.01$ for five of the six targets (4NTZ, 7O51, 7SEZ, 7VNX, 1L63), and $\rho = 0.005$ for 8DWN, where the larger value produced visible backbone distortion at the early guided steps.

For the likelihood weights in Eq.~\ref{log_likeli}, we use $\lambda_{\text{gauss}} = 0.9$ and $\lambda_{\text{rice}} = 0.1$ for all six targets. We picked these weights empirically. The Rice likelihood is the formally correct noise model for amplitude-only crystallographic data, but at $t_g$ the intermediate $\mathbf{X}_t$ still carries substantial residual error from the prior, and we found the heavier-tailed Rice term occasionally drove unstable gradient norms there. Down-weighting $L_{\text{rice}}$ and letting $L_{\text{gauss}}$ dominate smooths this out, while keeping enough Rice contribution to retain its calibration to amplitude noise. In our sweeps, any $\lambda_{\text{rice}} \in \{0,\,0.1,\,0.2\}$ paired with the corresponding much larger $\lambda_{\text{gauss}}$ produced comparable results; pushing $\lambda_{\text{rice}}$ higher than this consistently degraded the early-step trajectory.

For the post-DPS refinement objective we use the crystallographic $R$-factor,
\[
R \;=\; \frac{\sum_{\mathbf{h}}\big|\,|\mathbf{F}_{\text{o}}(\mathbf{h})| - |\mathbf{F}_{\text{c}}(\mathbf{h})|\,\big|}{\sum_{\mathbf{h}} |\mathbf{F}_{\text{o}}(\mathbf{h})|},
\]
across all six targets. The loss used during the refinement stage differs from that of the guidance stage. During guided diffusion the intermediate $\mathbf{X}_t$ is still noisy and we need measurement likelihood gradients to drag it toward the posterior data manifold, which is what the $\lambda_{\text{gauss}} L_{\text{gauss}} + \lambda_{\text{rice}} L_{\text{rice}}$ combination provides. By the refinement stage the predicted structure is clean, and $R$ is the standard scalar figure of merit reported by the crystallographic community. Minimizing $R$ at that point directly minimizes the metric we report in Table~\ref{tab:exp_results}. We use $N_{\text{refine}} = 50$ steps for 4NTZ, 7VNX, and 1L63, and $N_{\text{refine}} = 100$ for 8DWN, 7O51, and 7SEZ, which took longer to converge on $R$. Beyond these values we saw $\Delta R < 10^{-3}$ per additional 20 steps and moreover degradation in geometry, so we treated them as a reasonable trade-off between runtime and final metrics.

Table~\ref{tab:hyperparams_supp} summarizes the choices across the six targets. Most hyperparameters are shared; only $\rho$ and $N_{\text{refine}}$ are adjusted per target.

\begin{table}[h]
\centering
\small
\caption{Hyperparameter choices for the six experimental targets used in Table~\ref{tab:exp_results}. Only $\rho$ and $N_{\text{refine}}$ vary across targets.}
\label{tab:hyperparams_supp}
\begin{tabular}{lccccccc}
\toprule
Target & $t_g$ & $\rho$ & $k_{\text{total}}$ & $k_{\text{mask}}$ & $(\lambda_{\text{gauss}}, \lambda_{\text{rice}})$ & Refine loss & $N_{\text{refine}}$ \\
\midrule
4NTZ & 50 & 0.01  & 0.5 & 0.35 & (0.9, 0.1) & $R$-factor & 50  \\
8DWN & 50 & 0.005 & 0.5 & 0.35 & (0.9, 0.1) & $R$-factor & 100 \\
7O51 & 50 & 0.01  & 0.5 & 0.35 & (0.9, 0.1) & $R$-factor & 100 \\
7SEZ & 50 & 0.01  & 0.5 & 0.35 & (0.9, 0.1) & $R$-factor & 100 \\
7VNX & 50 & 0.01  & 0.5 & 0.35 & (0.9, 0.1) & $R$-factor & 50  \\
1L63 & 50 & 0.01  & 0.5 & 0.35 & (0.9, 0.1) & $R$-factor & 50  \\
\bottomrule
\end{tabular}
\end{table}

\subsection{Broader impacts}

CrystalBoltz aims to make X-ray crystallography faster and more accessible by integrating experimental measurements directly into a generative structure model. The most immediate beneficiaries are crystallographers, who currently spend substantial time on manual refinement and intervention when prediction-only models fail to recover condition-specific conformations. By bringing the experimental signal into generative sampling and reducing per-target runtime from hours to minutes, our approach can shorten the gap between data collection and a reliable atomic model. This in turn supports downstream work in structure-based drug discovery, where identifying conformations that are functionally relevant but rare in the deposited record is often the bottleneck. The framework is also general: any structural-biology modality with a differentiable forward model fits the same recipe, suggesting a path toward a unified treatment of experimental conditioning across X-ray, cryo-EM, and related techniques.

Two practical considerations follow from this work. First, the method is intended to accelerate refinement rather than to remove human review from the workflow; the guidance and refinement stages are exposed as separate, configurable steps so that practitioners can inspect intermediate structures before committing to a deposited model. Second, if generative refinement becomes routine, model outputs may eventually re-enter training corpora through public archives such as the PDB. We expect this to motivate explicit provenance metadata for AI-assisted depositions, both to keep future structural priors grounded in experimental signal and to support reproducibility of refinement pipelines that depend on a fixed prior.

\end{document}